\documentclass[10pt,journal,compsoc]{IEEEtran}

\usepackage{times}
\usepackage{epsfig}
\usepackage{graphicx}
\usepackage{amsmath}
\usepackage{amsthm}
\usepackage{amssymb}
\usepackage{dsfont}
\usepackage{algorithmic}
\usepackage{color}
\usepackage{soul}
\usepackage{caption}

\usepackage{algorithm,algorithmic}
\usepackage{subcaption}
\usepackage{overpic}

\usepackage{siunitx}
\usepackage{booktabs}

\usepackage[pagebackref=false,breaklinks=true,letterpaper=true,colorlinks,bookmarks=false]{hyperref}
\newif\ifdraft
\drafttrue

\definecolor{orange}{rgb}{1,0.5,0}

\ifdraft

\else

\fi

\newcommand{\bL}[0]{\mathcal{L}}

\newcommand{\bx}[0]{\mathbf{x}}

\newcommand{\by}[0]{\mathbf{y}}

\newcommand{\real}{\mathbb{R}}

\newcommand{\Axons}[0]{\textbf{Axons}}

\newcommand{\Roads}[0]{\textbf{Roads}}

\newcommand{\OURS}[0]{\textbf{OURS}}
\newcommand{\RoadTracer}[0]{\textbf{RoadTracer}}
\newcommand{\QMIP}[0]{\textbf{QMIP}}
\newcommand{\Seg}[0]{\textbf{Segmentation}}

\newcommand{\parag}[1]{\vspace{0.3mm}\noindent{\bf #1}}

\begin{document}

\author{Agata~Mosinska,
			Mateusz~Kozinski,
			Pascal~Fua \textit{Fellow,~IEEE}
	\IEEEcompsocitemizethanks{
		\IEEEcompsocthanksitem A.Mosinska, M.Kozinski and P. Fua are with the Computer Vision Laboratory, \'{E}cole Polytechnique F\'{e}d\'{e}rale de Lausanne, Switzerland.
	
}
}

\title{Joint Segmentation and Path Classification of Curvilinear Structures}	

\IEEEtitleabstractindextext{
	
	\begin{abstract}
Detection of curvilinear structures in images has long been of interest. One of the most challenging aspects of this problem is inferring the graph representation of the curvilinear network. Most existing delineation approaches first perform binary segmentation of the image and then refine it using either a set of hand-designed heuristics or a separate classifier that assigns likelihood to paths extracted from the pixel-wise prediction. In our work, we bridge the gap between segmentation and path classification by training a deep network that performs those two tasks simultaneously. We show that this approach is beneficial because it enforces consistency across the whole processing pipeline. We apply our approach on roads and neurons datasets.
\end{abstract}

}

\maketitle

\section{Introduction}\label{sec:intro}

\begin{figure}
  \centering
  \begin{tabular}{cc}
   \includegraphics[width=0.49\linewidth]{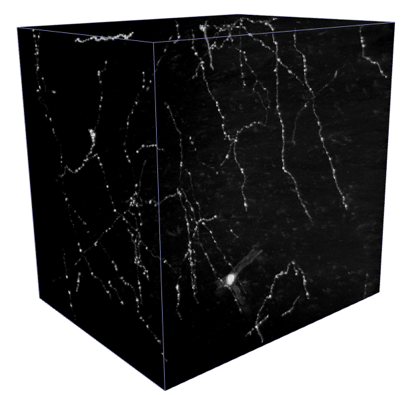} & 
    \includegraphics[width=0.49\linewidth]{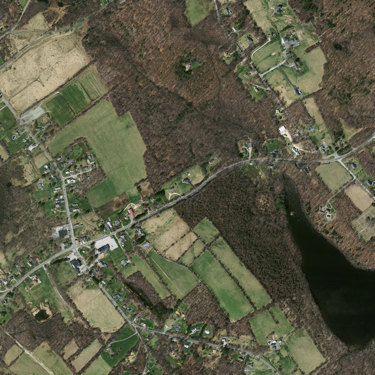} \\
    (a) & (b) 
  \end{tabular}
  \vspace{-3mm}
  \caption{{\bf Curvilinear structures.} (a) Axons and dendrites in a 2-photon light-microscopy stack. (b) Road network in aerial image.}
  \label{fig:teaser}
\end{figure}

\IEEEPARstart{A}{automated} delineation of curvilinear structures, such as those shown in Fig.~\ref{fig:teaser}, has been investigated since the inception of the field of Computer Vision in the 1960s and 1970s. Nevertheless, despite decades of sustained effort, full automation remains elusive when the image data is noisy and the structures are complex. As in many other fields, the advent of Machine Learning techniques in general, and Deep Learning in particular, has produced substantial advances, in large part because learning features from the data makes them more robust to appearance variations~\cite{Ganin14,Mnih12,Sironi16a,Wegner13,Turetken16a}. 

However, there remains a disconnect between two broad classes of techniques.  Some attempt to produce {\it segmentation masks} in which pixels potentially belonging to structures of interest are labeled as foreground and others as background. Others take such masks as input, usually along with a scalar image that indicates how confident the system is about the assigned labels, and attempt to produce a {\it delineation}, that is, a graph-like representation of the linear structures. Our own earlier work~\cite{Turetken16a} is representative of this dichotomy. We use one kind of classifier~\cite{Sironi16a} to produce the segmentation masks and another to score potential edges in the graph of which the desired delineation should be a subgraph. More recent Deep Learning-based approaches~\cite{Mattyus17,Ventura18} also rely on two independently-trained classifiers, one to perform segmentation and the other to classify potential connections.

In this paper, we bridge the gap between these two key steps so that they can be jointly optimized. To this end, we train a network made of an encoder and two different decoders to perform two separate tasks. First, given only an image,  produce a segmentation mask. Second, given a path between two points as an additional input,  return the likelihood that this path corresponds to a linear structure really present in that image. This  enables us to compute a segmentation, build a graph whose edges correspond to candidate linear structures, weigh them, and only retain the best ones. This is a much streamlined version of our earlier approach~\cite{Turetken16a}. Yet, because it enforces consistency throughout the processing chain, it performs better.

Our contribution is therefore a unified approach to segmenting linear structures and classifying linear paths. The intuition behind it is that these two tasks are closely related and should rely on the same features, specifically those that the decoder part of our network produces. This results in a general-purpose approach. We will demonstrate that it outperforms state-of-the-art ones in a number of image modalities, including satellite and light microscopy images.


\section{Related Work}
\label{sec:related}

Most delineation algorithms start by finding where the curvilinear paths are in the image. We refer to this as segmentation and it typically amounts to labeling individual pixels as being part of the target structures or not. In many applications such as neuroscience or cartography, this represents a good first step but does not suffice. What really matters is to infer the connectivity of the graph these linear structures form. In this section, we discuss existing ways to address these two tasks that are always treated as being separate, which is what this work intends to change.

\subsection{Segmentation}
\label{sec:segm}

Detecting linear structures can be achieved using either hand-crafted or learned features. Optimally Oriented Flux (OOF)~\cite{Law08} and Multi-Dimensional Oriented Flux (MDOF)~\cite{Turetken13c}, its extension to irregular structures, are successful examples of the former. Their great strength is that they do not require training data, but this advantage comes at a cost. Both techniques are prone to failure in more challenging cases, especially when input images feature a wide variation of scales or appearances, or when the imaged structures are highly irregular. In such situations, learning-based methods have an edge and several approaches have been proposed over the years. For example, Haar wavelets~\cite{Zhou12} or spectral features~\cite{Huang09} have been used as input to a classifier. In~\cite{Sironi16a}, the classifier is replaced by a regressor that predicts the distance to the closest centerline, which enables estimating the width of the structures.

As in many other areas of computer vision, many of these early learning-based methods have now been superseded by ones that rely on deep networks. For road delineation purposes, this was first done in~\cite{Mnih10}, directly using image patches as input to a  fully connected neural net. While the patch provided some context around the linear structures, it was still relatively small due to memory limitations. With the advent of Convolutional Neural Networks (CNNs), it became possible to use larger receptive fields. In~\cite{Ganin14}, CNNs were used to extract features that could then be matched against \emph{words} in a learned dictionary. The final prediction was made based on votes from nearest neighbors in the feature space. A fully-connected network of ~\cite{Mnih10} was replaced by a CNN in~\cite{Mnih13} for road detection. In~\cite{Mattyus17}, a differentiable Intersection-over-Union loss was introduced to obtain a road segmentation, which was then used to extract a graph of the road network. In the task of edge detection, nested, multiscale CNN features were utilized by Holistically-Nested Edge Detector~\cite{Xie15} to directly produce an edge map of an entire image.

In the biomedical field, the VGG network~\cite{Simonyan15}, pre-trained on real images, has been fine-tuned and augmented by specialized layers to extract retinal blood vessels~\cite{Maninis16}. Similarly the U-Net~\cite{Ronneberger15}, has been shown to give excellent results for biomedical image segmentation and is currently among the methods that yield the best results for neuron boundaries detection in the ISBI'12 challenge~\cite{arganda15}. Newer methods have focused on introducing features~\cite{Ganin15,Ghafoorian18,Singh18} and architectures~\cite{Cheng17,Xie15} better suited for this task. In our work~\cite{Mosinska18}, we have introduced a new loss term to capture higher-level properties of linear structures, such as smoothness or continuity. 


\begin{figure}
  \centering
  \begin{tabular}{cc}
     \includegraphics[width=0.4\linewidth]{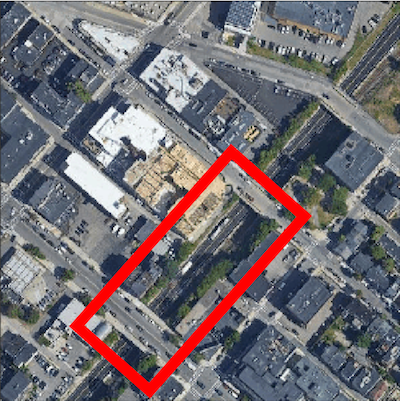} &
      \includegraphics[width=0.4\linewidth]{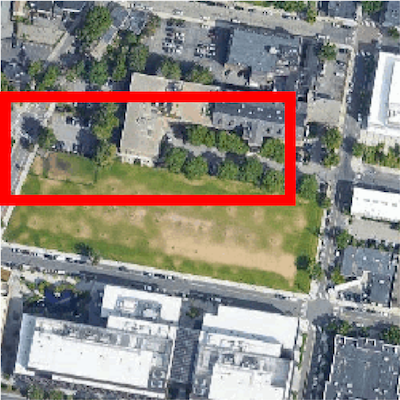} \\
   \includegraphics[width=0.4\linewidth]{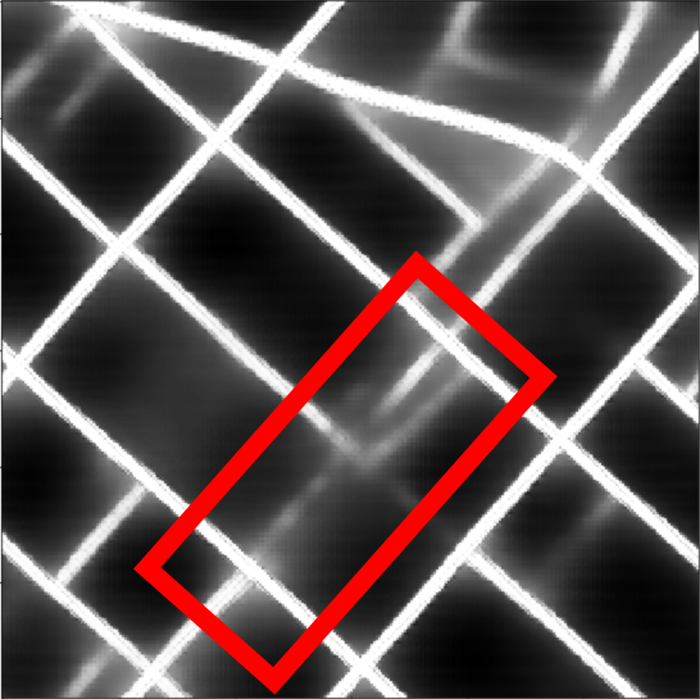} & 
    \includegraphics[width=0.4\linewidth]{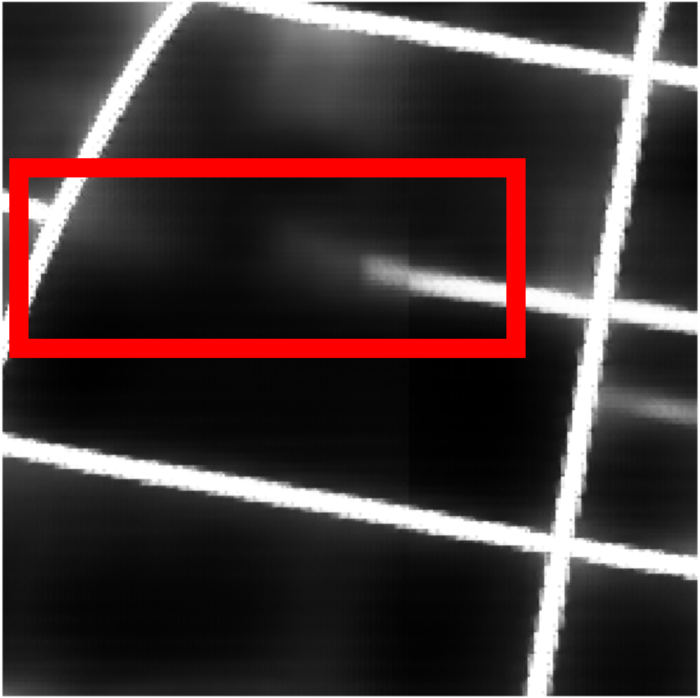} \\
    (a) & (b)
  \end{tabular}
  \vspace{-3mm}
  \caption{{\bf Ambiguities.} Two aerial images and the corresponding segmentations. In both cases, the red box denotes a part of the image in which the segmentation exhibits a gap. (a) The gap is not warranted as the street extends from one side of the box to the other. (b) The gap is legitimate because the path on the right side of the box is a driveways that stops at the entrance of a building.} 
  \vspace{-3mm}
  \label{fig:ambig}
\end{figure}

\subsection{Delineation}
\label{sec:delin}

The simplest way to infer connectivity information from a pixel probability mask is thresholding and skeletonizing it, followed by applying a set of heuristics to bridge the gaps and remove false detections\cite{Cheng17}. However, the choice of hand-designed heuristics is problem-specific, often requires setting many parameters, and usually  tackles only a small subset of the multitude of issues that can arise. 

Fig.~\ref{fig:ambig} depicts some of these difficulties that are best handled using a Machine Learning (ML) approach.  Most existing approaches to doing this comprise three steps. 

\begin{enumerate}

 \item Use one of the techniques discussed in Section~\ref{sec:segm} to assign a pixel of voxel a probability to belong to a linear or tubular structure, which we will refer to as a {\it tubularity map}. For example, in our earlier work, we used either OOF~\cite{Law08,Turetken13c} or a decision tree-based classifier~\cite{Sironi16a}. 
 
  \item Define a graph whose edges connect spatial locations and correspond to potential fragments of the target structures, for example by finding shortest paths between image locations where the distances take into account the tubularity map~\cite{Breitenreicher13,Montoya-Zegarra14,Wegner15,Turetken16a}. The edges are often referred to as \textit{paths}.
  
  \item Find a subgraph within that graph whose edges correspond to the candidate target structure fragments that are retained in the final delineation. The simplest way to do this is to look for a minimum spanning tree~\cite{Turetken12,Peng11b,Yang18c}, which can then be pruned. A more sophisticated approach that makes it possible to impose global geometric and topological constraints on the final delineation and allow loopy structures is to formulate the search for the optimal subgraph as Linear or Quadratic program~\cite{Turetken16a,Mosinska17}. A recent trend is to also use two separate deep networks for this purpose. One is trained to perform the segmentation and the other to score proposed paths~\cite{Mattyus17,Zhou18,Ventura18} or to trace the structure pixel by pixel and detect end points~\cite{Bastani18}.
   
\end{enumerate}

Note that even if both segmentation and delineation steps are accomplished using deep networks they remain distinct and are not trained to work together. This is the issue we address in this paper.

\section{Method}
\label{sec:method}


\begin{figure*}
  \centering
  \begin{overpic}[width=0.6\linewidth]{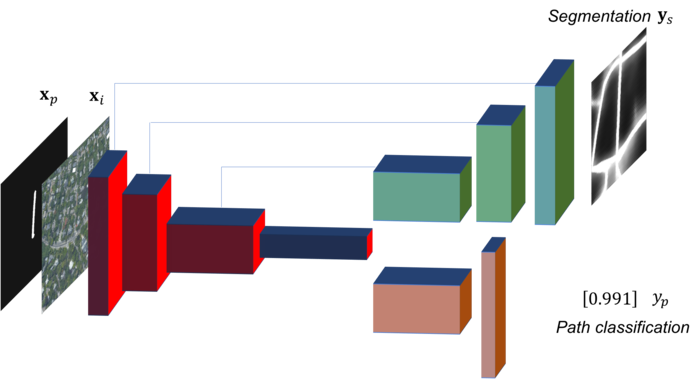}
    \put(12,-3){\makebox(40,5){\upbracefill}}
    \put(31,-4){$f_\mathrm{enc}$}
    \put(53,-3){\makebox(20,5){\upbracefill}}
    \put(62,-4){$f_p$}
    \put(53,44){\makebox(29,5){\downbracefill}}
    \put(66,48){$f_s$}
  \end{overpic}
  \vspace{5mm}
  \caption{{\bf Two-stream architecture.} Both branches share the same U-Net encoder that takes as input the image and a binary mask representing a candidate path. The first branch uses the U-Net decoder and skip connections to produce a tubularity map. The second branch relies on a simpler network to yield a classification score for the path.} 
  \label{fig:architecture}
\end{figure*}


\begin{figure*}
  \centering
  \begin{tabular}{ccccc}
  \includegraphics[width=0.19\linewidth]{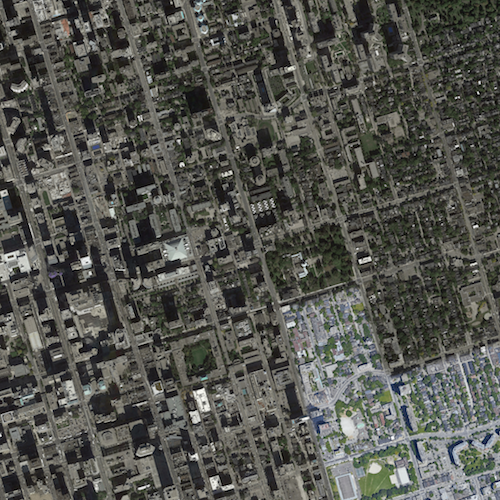} & 
  \includegraphics[width=0.19\linewidth]{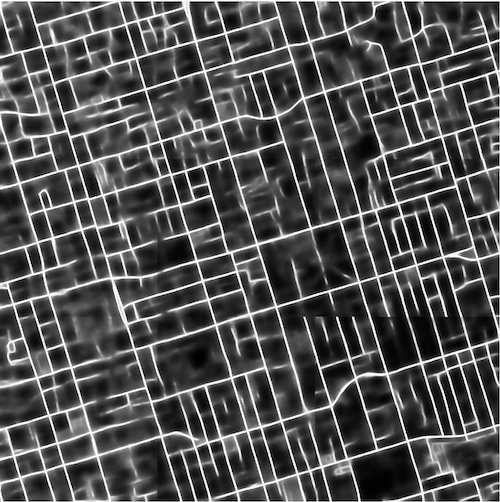} & 
   \includegraphics[width=0.19\linewidth]{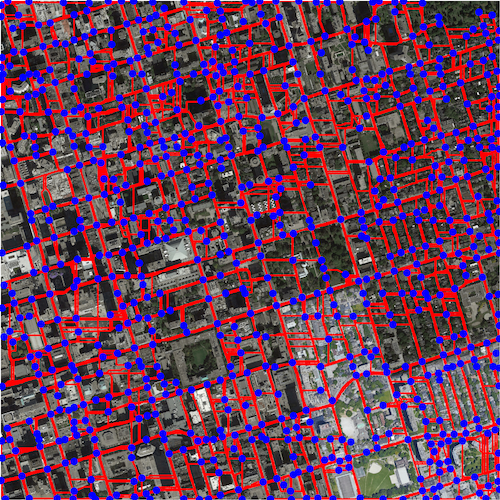} & 
    \includegraphics[width=0.19\linewidth]{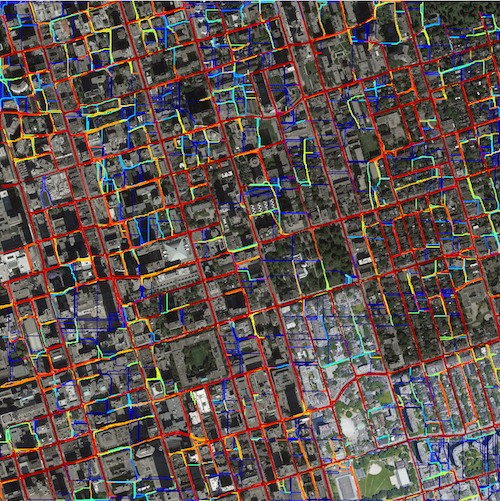} & 
    \includegraphics[width=0.19\linewidth]{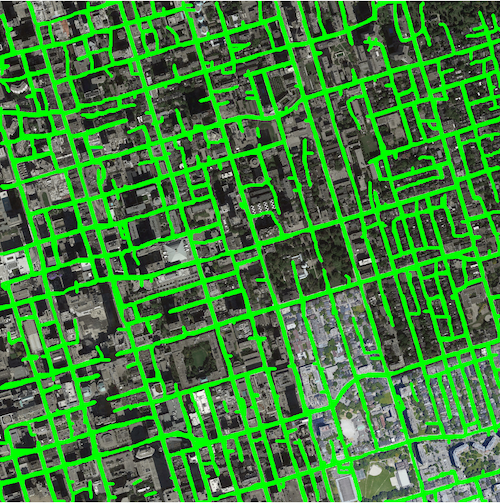} \\
    Image &  Tubularity map & Overcomplete graph & Scored paths & Result 
  \end{tabular}
  \vspace{-2mm}
  \caption{{\bf Delineation pipeline.} Given an image, we first compute a pixel-wise probability map of linear structures, which we also call a tubularity map. Next, we build a graph by finding nodes, shown as blue dots, and connecting them by shortest paths, shown in red. This graph is overcomplete in the sense that it contains the true linear structures and false positive linear structure candidates. To eliminate them, we run a classifier for each edge of the graph and remove the low-scoring edges.}
   \label{fig:pipeline}
\end{figure*}

Our approach belongs in the same class as the 3-step approaches described in Section~\ref{sec:delin}. We compute a tubularity map, use it to build an overcomplete graph whose edges correspond to candidate linear structures, assign each one a score, and only retain the best ones. Fig.~\ref{fig:pipeline} depicts these steps. However, unlike in~\cite{Turetken16a}, we did not find it necessary to perform a sophisticated linear and quadratic optimization to select an optimal subgraph. This is because we use the same network to compute the tubularity map and the edge weights, which ensures consistency and increases performance to the point where the more sophisticated approach to building the final delineation~\cite{Turetken16a}  is no longer needed to outperform the state-of-the-art. In this section we describe the architecture of this network, along with the associated training and testing procedures.

\subsection{Formalization}
\label{sec:formal}

For simplicity's sake, we formalize our approach in the context of 2D image delineation but it naturally extends to 3D image stacks. 

Let $\bx_{i} \in \real^{H\cdot{}W\cdot{}C}$ be a $W\times{}H$ $C$-channel image and $\bx_{p} \in \real^{H\cdot{}W}$  be a $W\times{}H$ binary image representing a candidate path, where $1$s denote the pixels belonging to the path. They are inputs to our encoder network $f_\mathrm{enc}$ and are depicted on the left side of Fig.~\ref{fig:architecture}. As shown on the right side of the figure, the latent representation $h=f_\mathrm{enc}(\bx_i,\bx_p)$ is fed to a decoder $f_s$, outputting a segmentation image $\by_s=f_s(h)$, and to another decoder $f_p$, outputting the path classification score $y_p=f_p(h)$.

For every pixel $q$ of the image $\bx_i$, the value of $\by_s[q]$ is understood to be the probability of assigning to $q$ the label $1$, that is, of belonging to a linear structure. Given the ground-truth $W\times{}H$ binary segmentation image $\by_s^{gt}$, we therefore want $\bL_{seg}$, the binary cross entropy summed over $\by_s$ and $\by_s^{gt}$ to be as small as possible. Similarly the classification score $y_p$ is taken to be the probability of the path being valid and we also want binary cross entropy $\bL_{p}$ given the ground-truth label to also be as small as possible. 

Therefore for each training example, 
we take the loss to be
\begin{equation}
\bL_{seg}(\by_s,\by_s^{gt}) + \eta_p \bL_{p}(y_p,y^{gt}_p)   \; ,
\label{eq:loss}
\end{equation}
where $\eta_p$ is a constant that ensures that the gradients associated to both losses have similar magnitudes. Both $\bL_{seg}$ and $\bL_{p}$ are computed using binary cross-entropy. Optionally, we can add a topology loss of~\cite{Mosinska18} to $\bL_{seg}$  to ensure that the global statistical properties of the segmentation are similar to those of the ground-truth.

\subsection{Architecture}
\label{sec:architecture}

To implement the network $f$ described above, we start from the U-Net architecture~\cite{Ronneberger15}. It is fully convolutional and comprises an encoder and decoder with skip connections. It has been shown to be state-of-the-art for binary segmentation of linear structures~\cite{Cicek16,Mosinska18}. To also perform path classification, we modified it to accept as input the binary mask $\bx_p$ in addition to the image $\bx_i$, as shown on the right side of Fig.~\ref{fig:architecture}. We then add a second branch depicted at the bottom of the figure. It connects to the encoder and outputs the classification score $y_p$ while the first branch still produces the segmentation probability map $\by_s$. Both are depicted on the right side of Fig.~\ref{fig:architecture}. 

More precisely the segmentation branch follows the standard U-Net design with four max pooling operations and skip connections between corresponding encoder and decoder layers. At each downsampling step the number of filters is increased by a factor two and the reverse occurs while upsampling. The classification branch uses the same encoder but a much simpler decoder that comprises an additional convolutional layer followed by max pooling and two fully connected layers. To speed-up convergence at training time, we use batch normalization. We also use current batch statistics at test time~\cite{Cicek16}. For 3D datasets we use the encoder with two max-pooling operations and two fully-connected path classification layers due to memory constraints.

In practice, to compute $\by_s$ and $y_p$, we perform a first pass through the network with $\bx_p$ replaced by an image entirely made of zeros, so that the result is only conditioned on the image. We then make a second pass using $\bx_p$ to obtain a value of $y_p$ that is conditioned both on the image and the input path. This guarantees that the same intermediate features encode both the segmentation map and the classification score. 


\begin{figure*}
  \centering
  \begin{tabular}{cc}
     \includegraphics[width=0.18\linewidth]{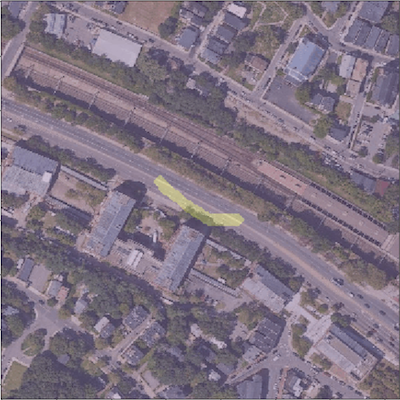} 
      \includegraphics[width=0.18\linewidth]{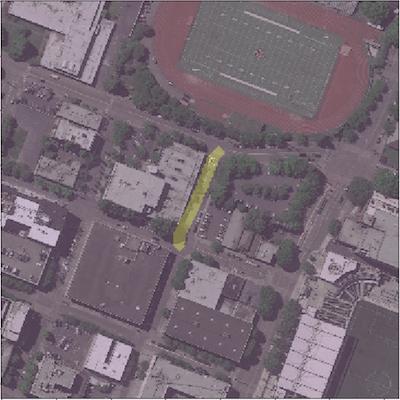} &
   \includegraphics[width=0.18\linewidth]{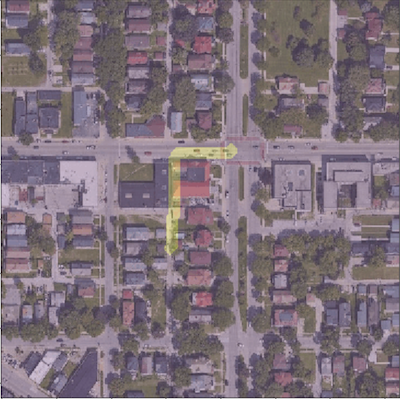}  
       \includegraphics[width=0.18\linewidth]{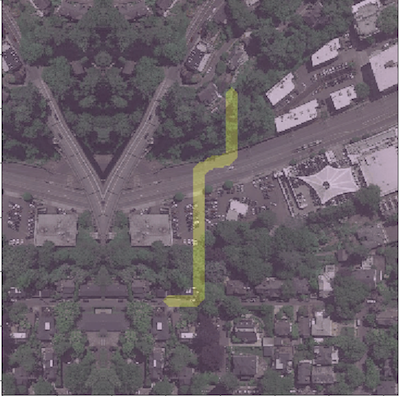}  \\
       
     \includegraphics[width=0.18\linewidth]{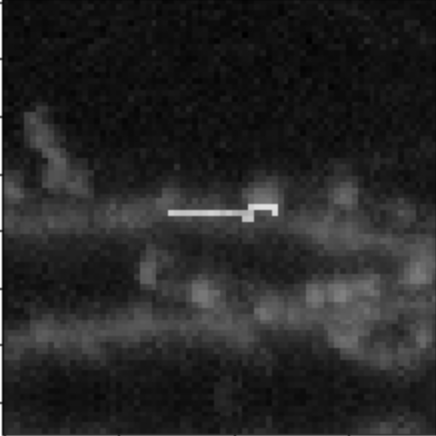} 
      \includegraphics[width=0.18\linewidth]{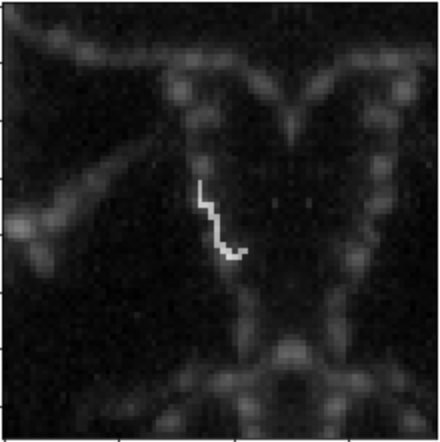} &
   \includegraphics[width=0.18\linewidth]{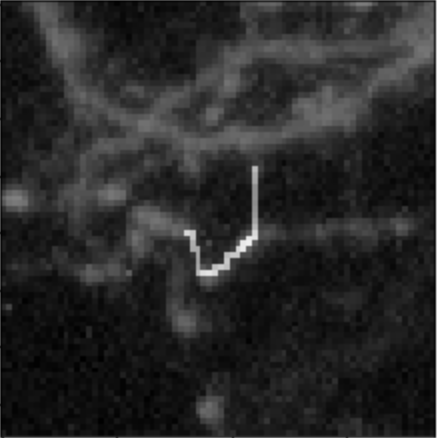}  
       \includegraphics[width=0.18\linewidth]{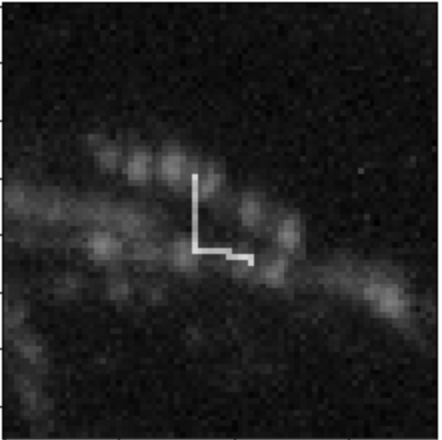}  \\       
       
    Positive samples & Negative samples  
  \end{tabular}
  \caption{{\bf Candidate road and neuron paths.} {\bf (Top row)} Road images with candidates overlaid in yellow.  {\bf (Bottom row)} 2-photon images with candidates overlaid in white. In both cases, we treat the paths that remain with linear structures as positive, even if they do not exactly follow the centerline. By contrast, paths that cross from one structure to another or take shortcuts as treated as negative.}
  \label{fig:paths}
\end{figure*}

\begin{figure}
  \centering
   \includegraphics[width=0.6\linewidth]{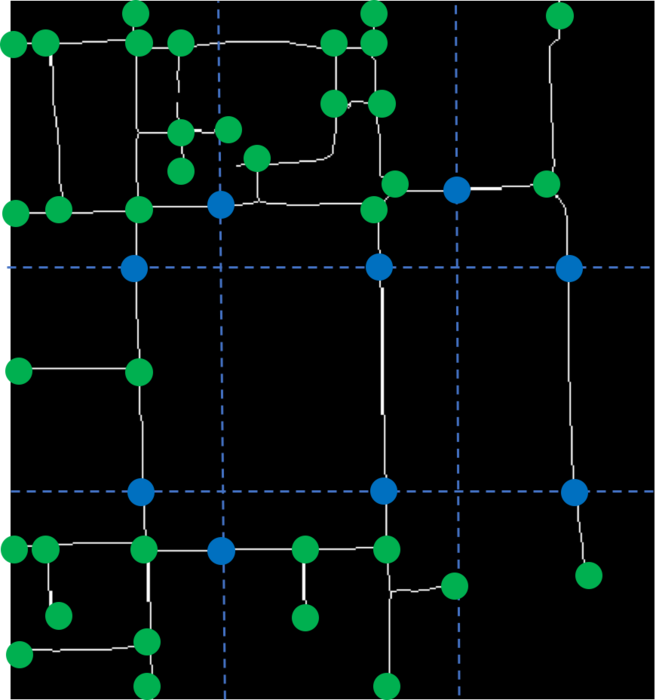} 
  \caption{\textbf{Creating graph nodes.} We skeletonize the ground-truth segmentation or tubularity map and find the topologically significant points, that is, the intersections and end-points shown as green disks. We use them as nodes of our over-complete graph. As they can be far from each other,  we additionally introduce the regular grid depicted by the dashed lines and use its intersections with the tubularity skeleton, shown as blue disks, as additional nodes. To prevent nodes from being too close from each other, we introduce an exclusion zone of radius $\varepsilon$, which is also the radius of the green disks, in which no additional nodes can be added. This approach guarantees that the distance between nodes that ought to be connected is always between $\varepsilon$ and $d$.}
  \label{fig:samples}
\end{figure}

\subsection{Training}
\label{sec:training}

To train our network, we need pairs of images $(\bx_i,\by_s^{gt})$ and candidate paths $\bx_p$ that are labeled as being either valid or not. As will be discussed in Section~\ref{sec:results}, there are publicly available training databases with ground truth segmentations we can use to obtain sufficiently large numbers of $\bx_i$ and $\by_s^{gt}$. Given those, a simple way to find path candidates would be to sample positives from the ground-truth segmentation and use random paths as negatives. However, this would produce candidates that are far too easy to classify to provide a useful supervisory signal. Instead, we initially train only the segmentation branch of our network. When the loss stabilizes, we use it to compute a tubularity image. We then use this image to create positive and negative path samples. This enables us to jointly train the two branches of the network. This way, the paths we use to train the path-scoring branch of our network are realistic ones and resemble those we are likely to encounter at inference time, such as those of Fig.~\ref{fig:paths}. Note that the true positives do not have to exactly follow the centerline as long as they remain along the ground-truth path.

Given a segmentation tubularity image, our path generating procedure performs the following three steps:

\begin{enumerate}

\item {\bf Finding graph nodes.} We threshold and skeletonize the tubularity image and find {\it significant points}, that is, intersections and endpoints. Endpoints have a single neighbor while intersections have more than two. As these significant points may be unevenly spaced in the image, we also sample the skeleton at regular intervals to limit the maximum distance between neighboring samples . 

\item {\bf Connecting graph nodes.} Fig.~\ref{fig:samples} depicts this node creation process, which guarantees that nodes that ought to be connected are within a distance $d$ of each other. To account for small inaccuracies, we connect all pairs of nodes that are closer than a certain distance. This yields an overcomplete graph such as the one depicted by Fig.~\ref{fig:pipeline} in which each edge corresponds to a shortest path between two nodes.  We use the A$^*$ algorithm to extract these paths. Given a start and a target node, it iteratively grows the path by adding to it the pixel $(x,y)$ that neighbors the current path endpoint and minimizes
\begin{align}
f(x,y)  &= c(x,y) + h(x,y) \; , \\
c(x,y) & = 1.1 - p(x,y)  \; , \\
h(x,y) & = 0.5*d(x,y) \; ,
\end{align}
where $p$ is the tubularity value returned by our classifier and $d$ is the Euclidean distance to the target. In effect, $c$ is the cost of adding a new pixel to the path and $h$ approximates the cost of the shortest path from $(x,y)$ to the target. We found this to deliver a good compromise between computation speed on potentially very large graphs and close approximation of the true shortest path.

\item {\bf Selecting positive and negative samples.}  Paths that overlap with segmentation ground truth by more than 90\% are considered as positive. The others are assigned a negative label. This yields positive and negative samples as those depicted by Fig.~\ref{fig:paths}. This heuristic treats as positive the paths that remain within the linear structures, even if they do not exactly follow its centerline. By contrast, those that cross from one structure to another  using a shortcut are labeled as negative. In earlier work~\cite{Turetken16a}, we found that the latter were one of the main sources of errors in the final network topology and that it was crucial to eliminate them.

\end{enumerate}

To keep the notations simple in Section~\ref{sec:formal}, we formulated our approach in terms of whole images. In practice, our network operates on patches that are cropped from the training images. To select such patches for training purposes, we use the positive and negative paths selected as discussed above and crop the image around them. If a path does not fit within the receptive field of our U-Net encoder, we split it.

\subsection{Inference}

Inference proceeds in a way that closely mirrors training, as shown in Fig.~\ref{fig:pipeline}. We use the segmentation branch of our now trained two-stream U-Net to compute the tubularity map. We then skeletonize it and create the graph nodes using the procedure depicted  by Fig.~\ref{fig:samples}. We connect them using the same A$^*$ algorithm as before and score the resulting paths using the second branch of our two-stream U-Net. Finally, we retain only the highest scoring ones. 

The key to the effectiveness of this procedure is that we use the {\it same} network to compute the tubularity map and to score the paths, which guarantees that both operations rely on the same image features.

\section{Results}
\label{sec:results}

\subsection{Datasets and Baselines}

We evaluate our approach on two datasets depicting very different curvilinear structures: 

\begin{itemize}

\item \Roads{}~\cite{Bastani18}: One of the biggest publicly available road network datasets, featuring highways, urban roads, and rural paths. The training set comprises images of 25 cities and the training set 15 {\it other} cities. The images come from Google Maps,  are of size 4096$\times$ 4096, and each covers area of around 20~km$^2$. The ground-truth was obtained from Open Street Maps and may therefore  be noisy. Since the training and testing sets come from different cities,  the database is well-suited to assess generalization abilities.  Due to computational constraints we downsample the original images and ground-truth by the factor of two. In order to extract overcomplete graph we further downsample the tubularity map by half. However, the evaluation metrics are computed and compared in the original resolution by upsampling the results.

\item \Axons{}: 3D stacks of two-photon laser scanning microscopy images depicting axons in mouse brains at different stages of their development. We use 8 volumes of size ranging from 217$\times$206$\times$54 to 322$\times$277$\times$136 for training and two other volumes of size 500$\times$350$\times$200 and resolution $0.27\times0.27\times0.9$\si{\micro\meter} for testing.

\end{itemize}

We set the uniform node sampling distance $d$ equal to 250 for \Roads{} and 30 for \Axons{} and the maximum connection distance between two nodes to $1.5d$ for \Axons{} and $1.1d$ for \Roads{}.

We will refer to our method as  \OURS{} and compare it to two state-of-the-art graph-based approaches for road and neuron delineation:

\begin{itemize}

\item \RoadTracer{}~\cite{Bastani18}: A recent method for road detection based on a Convolutional Neural Network trained to decide the direction of the next pixel on the traced path. We use the results reported by the authors on the \Roads{} dataset for evaluation purposes.

\item \QMIP{}~\cite{Turetken16a}: Our previous approach to delineation that follows similar steps as \OURS{}, but requires two separate classifiers - one to compute tubularity and the other to classify paths. It also performs Mixed-Integer Programming optimization to find the optimal subgraph.

\end{itemize}

We also compare our results to those obtained by simply skeletonizing the segmentation, which we will refer to as \Seg{}.


\begin{figure*}
  \centering
  \begin{tabular}{cc}
   \includegraphics[width=0.4\linewidth]{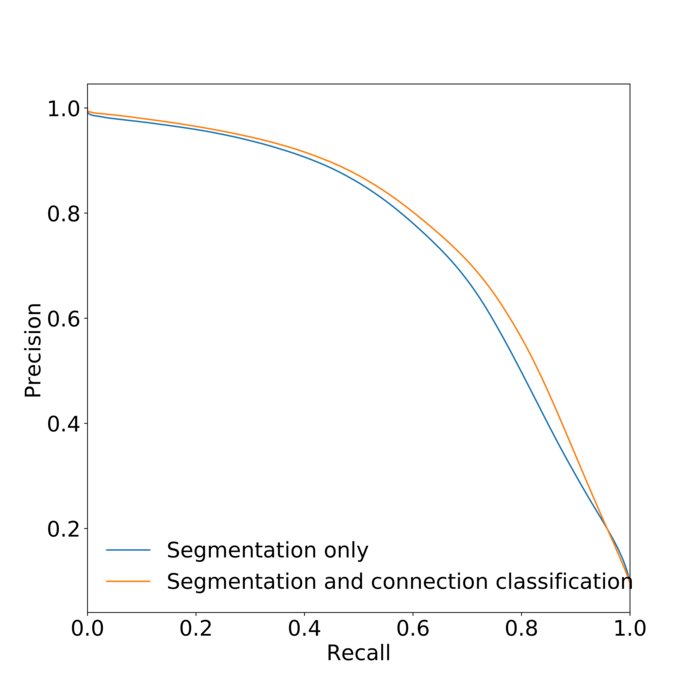} & 
    \includegraphics[width=0.4\linewidth]{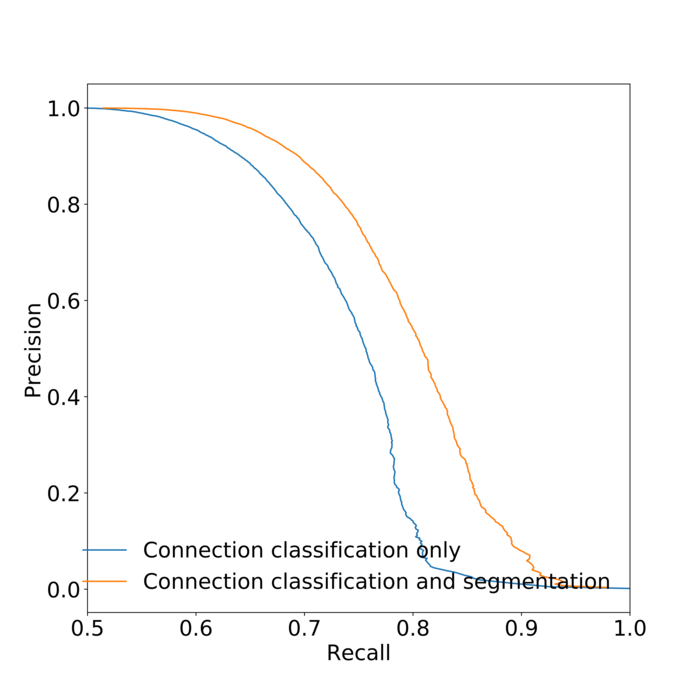} \\
    (a) Segmentation & (b) Path Classification 
  \end{tabular}
  \vspace{-2mm}
  \caption{{\bf Task sharing.} Learning both path classification and segmentation simultaneously boosts path classification, and does not harm segmentation, compared to training two networks seperately.}
  \label{fig:PR_curve}
\end{figure*}


\begin{figure*}
  \centering
  \begin{tabular}{ccc}
  \includegraphics[width=0.33\linewidth]{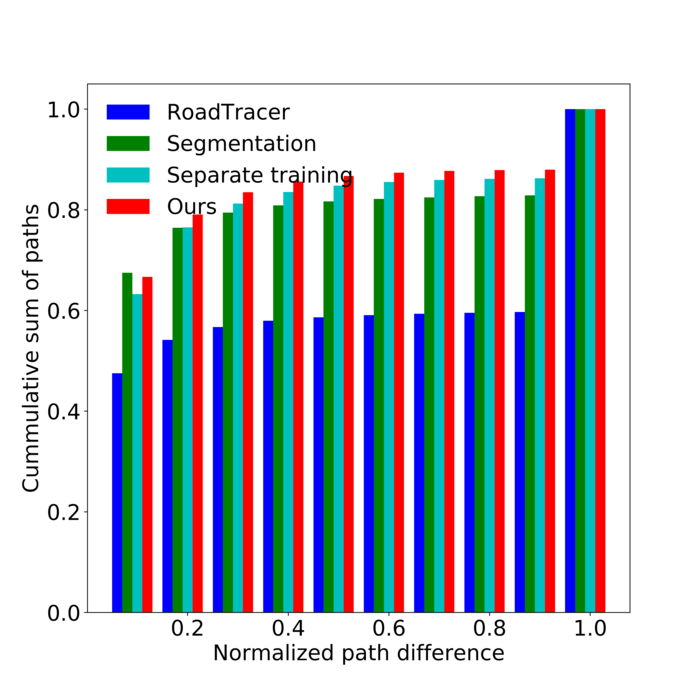} & 
   \includegraphics[width=0.33\linewidth]{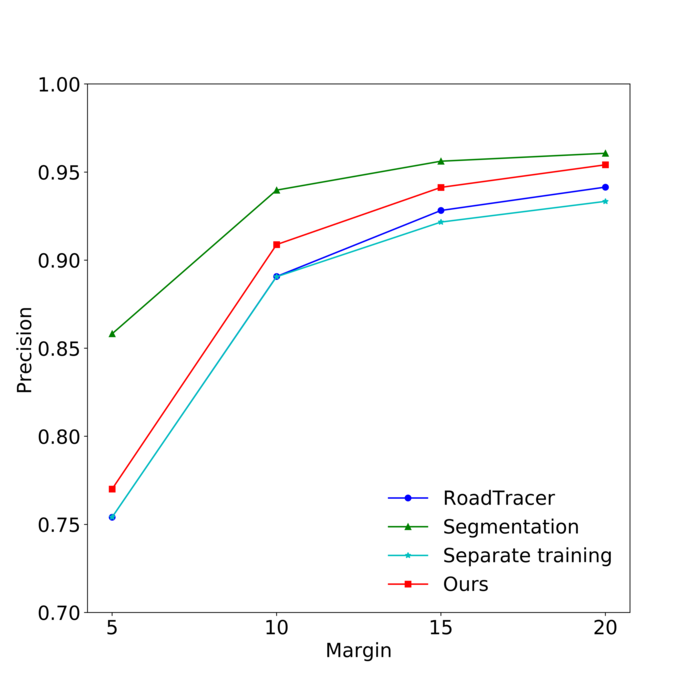} & 
    \includegraphics[width=0.33\linewidth]{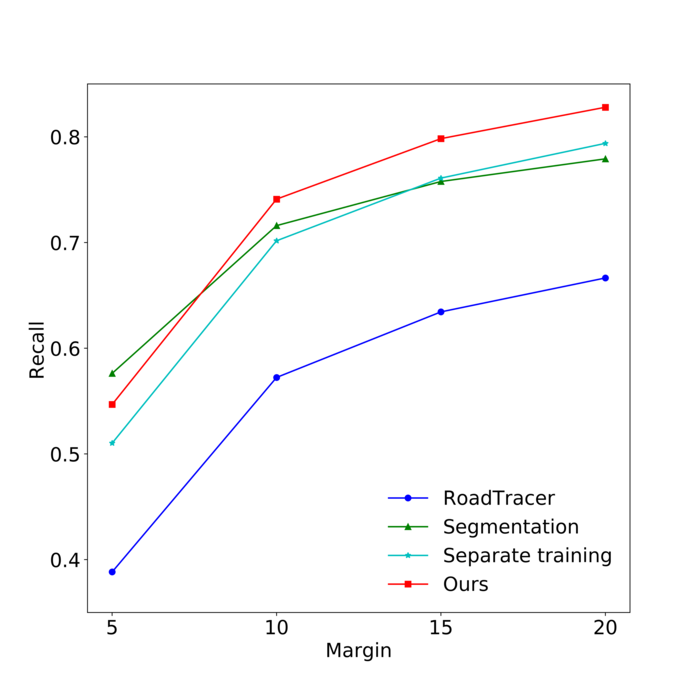} \\
    
      \includegraphics[width=0.33\linewidth]{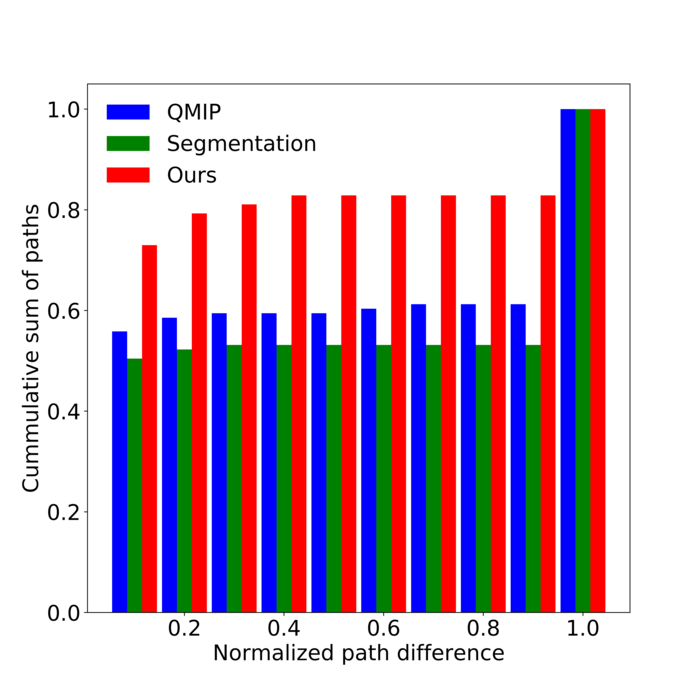} & 
   \includegraphics[width=0.33\linewidth]{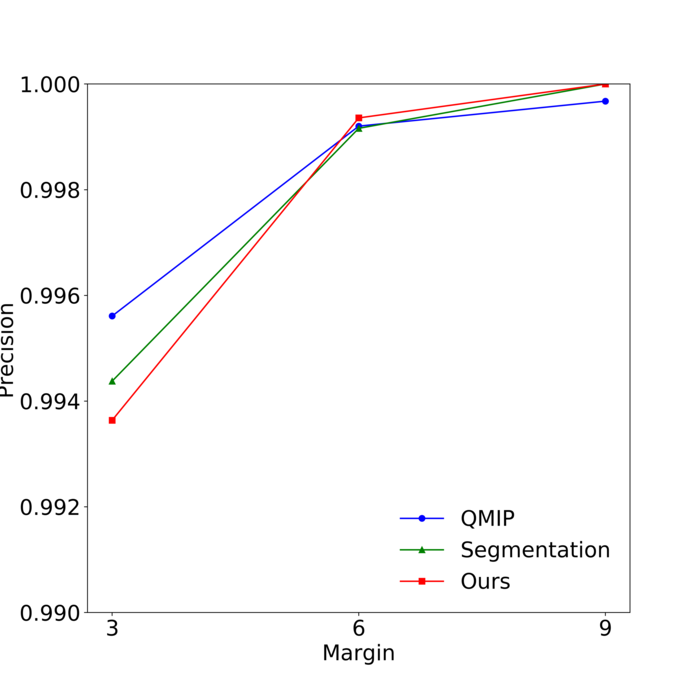} & 
    \includegraphics[width=0.33\linewidth]{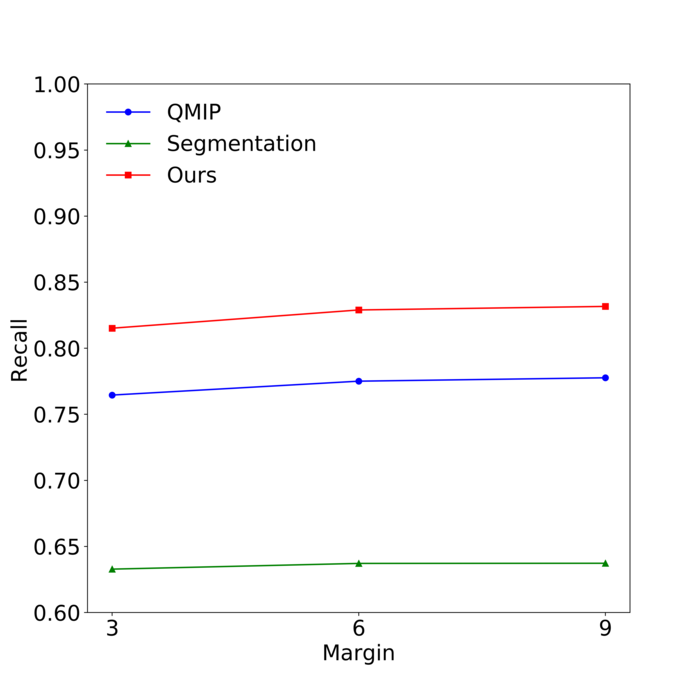} \\
    
    Normalized Path Difference &  Topological Precision & Topological Recall 
  \end{tabular}
  \caption{{\bf Quantitative results.} {\bf Top:} Roads. {\bf Bottom:} Axons. From left to right, cumulative distribution of the Normalized Path Distances; topological precision and recall as a function of the distance threshold $m$ (in pixels) used to compute them. NPD considers all connected pairs of significant points and it is therefore a global measure, while topological precision and recall computes shortest paths only between neighbouring significant points and as a result it is a local metric.}
  \label{fig:quant}
\end{figure*}

\subsection{Evaluation Metrics}
\label{sec:metrics}


\begin{figure*}
  \centering
  \begin{tabular}{ccc}
  \includegraphics[width=0.4\linewidth]{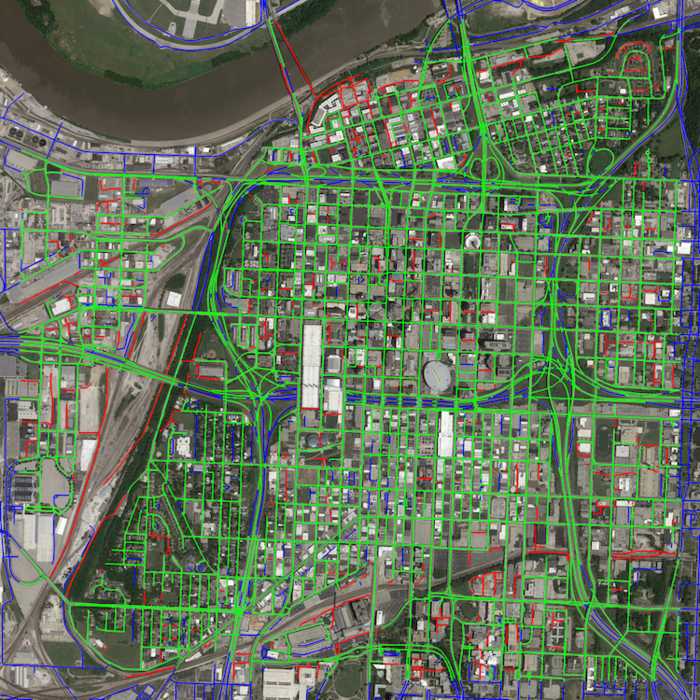} & 
   \includegraphics[width=0.4\linewidth]{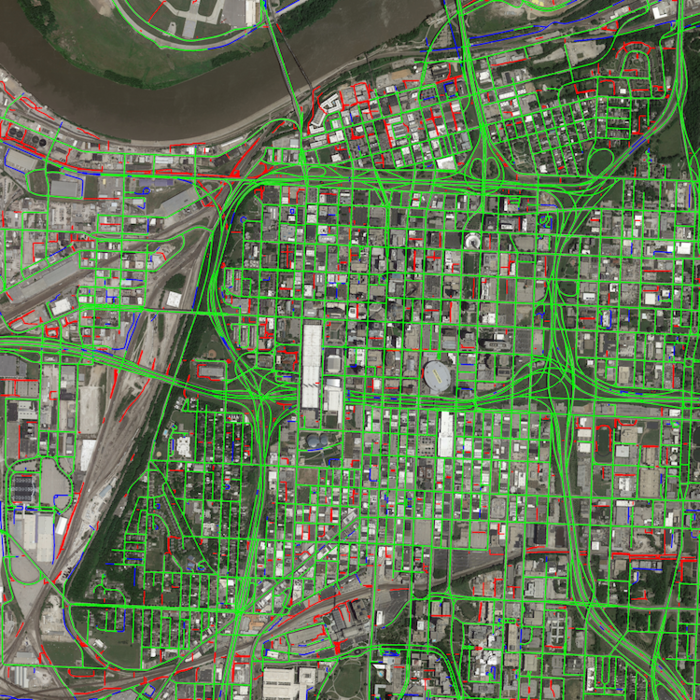} \\
   (a) & (b)
  \end{tabular}
  \caption{{\bf Roads qualitative results.} Kansas City road network (a) \RoadTracer{} (b) \OURS{}. True positive parts of the network are shown in green, false positive in red and false negative in blue. Matching is done within a radius of 10 pixels. \RoadTracer{} misses the roads at the periphery because the endpoints are not found as well as some of the smaller roads and intersections, which \OURS{} finds. }
  \label{fig:qualitative_roads}
\end{figure*}

\begin{figure*}
  \centering
  \begin{tabular}{ccc}
  \includegraphics[width=0.33\linewidth]{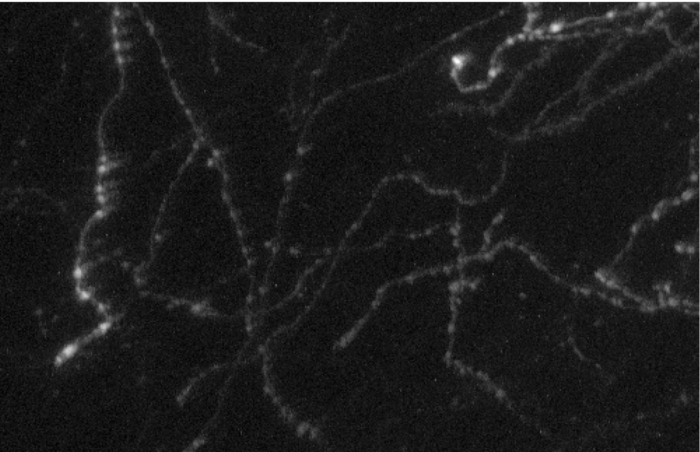} &
  \includegraphics[width=0.33\linewidth]{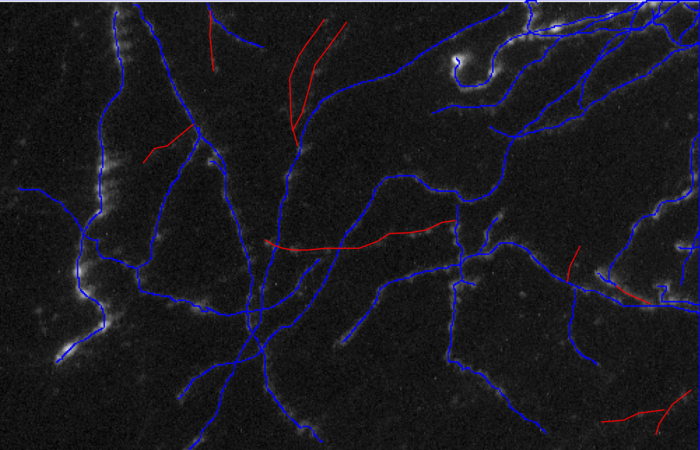} & 
   \includegraphics[width=0.33\linewidth]{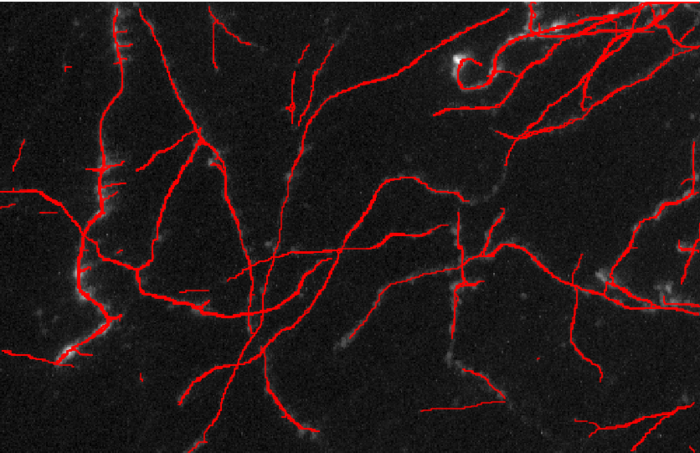} \\
   (a) & (b) & (c)
  \end{tabular}
  \vspace{-2mm}
  \caption{{\bf Axons qualitative results.} (a) maximum intensity projection of the image (b) overlaid ground-truth and (c) \OURS{} reconstruction of one of the \Axons{} test images. The blue ground-truth tracings were done by the annotator but the additional ground-truth tracings correspond to the neurons that can be still seen in the image.}
  \label{fig:qualitative_axons}
\end{figure*}

Our algorithm aims to produce paths that approximate the centerline of linear structures while preserving the overall topology of the network. We therefore rely on topology-aware metrics that have been proposed in the literature to assess the performance of delineation algorithms. They all rely on computing and comparing the shortest path between pairs of intersections or endpoints, which we defined as topologically significant points in Section~\ref{sec:training}. The significant points and shortest paths are computed in the ground-truth and then the corresponding significant points and shortest paths are found in the predicted graph. Where they differ is in the way the similarity between the two paths is measured and used. 

\parag{Normalized Path Difference~\cite{APLS}.}

Let $a^*$ and $b^*$ be the two significant points being connected by a ground-truth shortest-path of length $l^*$. If there are corresponding significant points $a$ and $b$ within a radius $R$ in the prediction, we pick the closest ones, find the length $l$ of the shortest-path connecting them, and compute the dissimilarity score
\begin{equation}
 \min\lbrace\frac{| l -l^*|}{l^*},1\rbrace \; .
 \label{eq:dissSore}
\end{equation} 

If there are no corresponding points in the prediction, the score is taken to be 1. We compute the normalized path difference for all connected pairs of ground-truth significant points. In practice we take the value of the radius $R$ to be 40 pixels for \Roads{} and 10 pixels in \Axons{}, which  is roughly the minimum spacing between two intersections. We then plot cumulative distribution of all dissimilarity scores.

\parag{Topological precision and recall.}

A variant of this measure was proposed in~\cite{Mattyus17}. However, the original definition, as the Normalized Path Difference, considers only lengths of the corresponding shortest paths and does not account for how closely the predicted shortest path follows the ground-truth one. We noticed, however, that two paths may have similar lengths but follow different routes. For that reason we match the corresponding shortest paths pixel-by-pixel. For all pairs of significant points that are neighbors in the ground-truth graph, we consider the path connecting them and the corresponding shortest-path in the predicted graph, as described above.  We can then write

\begin{equation}
\mbox{precision}    = \frac{1}{\sum{n_m}}\sum{\frac{n_m}{n_t}n_m}  \;
\mbox{and recall}   = \frac{\sum n_m^* }{\sum n_t^*}   \; ,
 \label{eq:precision}
\end{equation}

where $n_m$  is the number of pixels along the predicted path that are within distance $m$ of the ground-truth one and $n_t$ is the total number of pixels in the predicted path. Similarly,  $n_m^*$  is the number of pixels along the ground-truth path that are within distance $m$ of the predicted one and $n_t^*$ is the total number of pixels in the ground-truth path. When there is no corresponding path, $n_m$ and $n_m^*$ are set to zero. Precision captures how accurately the predicted path locations are and is weighted by the length of the path. Recall indicates the overall proportion of the ground truth that is effectively modeled by these paths. 

\subsection{Analysis and Comparisons}

We first studied the effect of joint segmentation and path classification training on how well both of these tasks are performed. While we found that it does not affect segmentation significantly, it considerably boosts the performance of path classification, as shown in Fig~\ref{fig:PR_curve}. This is likely because both tasks share similar features, which emphasizes the dependence of path classification and segmentation and helps distinguish negative paths, even if they partly coincide with a true road. Moreover, task-sharing also improves the results in terms of the final topology metrics, as shown at the top of Fig.~\ref{fig:quant}.

Fig.~\ref{fig:qualitative_roads} and Fig.~\ref{fig:qualitative_axons} depict some of our qualitative results on the \Roads{} and \Axons{} datasets.  In Fig.~\ref{fig:quant}, we report the corresponding quantitative results and compare them against those of our baselines. All three road extraction methods deliver similar precision (apart from small margin for \Roads{}, where \Seg{} has got better precision than other two methods), but our approach has got much higher recall. The histogram of normalized path differences also indicates that joint path classification and segmentation reduces the number of disconnected segments. 

\section{Conclusion}
\label{sec:conclusion}

We presented a method for joint segmentation and connection classification of curvilinear structures. We use the fact that those two tasks share similarities to train a network that can perform both tasks simultaneously. As a result they are optimized for each other and our approach outperforms state-of-the-art methods on very different datasets depicting roads and neurons.

\bibliographystyle{IEEEtran}
\bibliography{string,vision,biomed,learning,optim,misc}

\end{document}